# Deep one-gate per layer networks with skip connections are universal classifiers

Raul Rojas
Department of Mathematics and Statistics
University of Nevada Reno
October 2025

**Abstract**

This paper shows how a multilayer perceptron with two hidden layers, which has been designed to classify two classes of data points, can easily be transformed into a deep neural network with one-gate layers and skip connections.

## 1   Introduction

As shown in [1], deep one-gate per layer networks can perfectly separate points belonging to two classes in an *n*-dimensional space. Here, I present an alternative proof that may be easier to understand. This proof shows that classical neural networks that separate two classes can be transformed into deep one-gate-per-layer networks with skip connections.

A perceptron receives a vector input and divides input space into two subspaces: the positive and negative half-spaces (Fig. 1a). The perceptron's bias input can be provided by the input vector, which increases its dimensionality by one. The bias connection to the perceptron is assigned weight $n + 1$. The computation for each perceptron with weight vector $w^T = (w_1, w_2, \ldots, w_{n+1})$ involves comparing the inner product of the input vector $x$ and the weight vector $w$, to the threshold 0:

$$x_1 w_1 + x_2 w_2 + \cdots + x_n w_n + 1 \cdot w_{n+1} \geq 0.$$

The output is one if the inequality is true and zero otherwise.

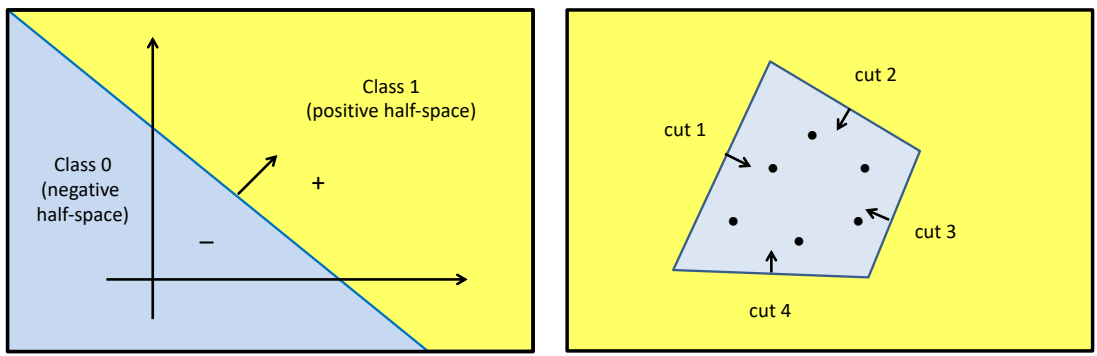

Figure 1: (a) A single perceptron divides input space into two halves. (b) Several perceptrons can define a convex region of input space. The union of the positive half-spaces of the four perceptrons delimits the gray region. This can represent a class of points in a classification problem, such as gray versus yellow points.

Perceptrons are well-known to be able to implement Boolean logic functions such as disjunction, conjunction, and negation for binary n-dimensional inputs by appropriately selecting the weights and bias attached to the perceptron input channels [2].

Assuming the classification problem involves two classes that can be separated by enclosing the points of one class within convex polytopes, then Fig. 1b shows one possible approach: if the grey class can be fully enclosed by several cuts, the intersection of the positive half-spaces (its logical conjunction) provides a separation of the two classes. If the class of grey points is found in several "islands" or clusters, we can separate each cluster from the background using different enclosing polytopes, as shown in Fig. 2. In that figure, nine cuts and three intersections of positive half-spaces provide the required classification.

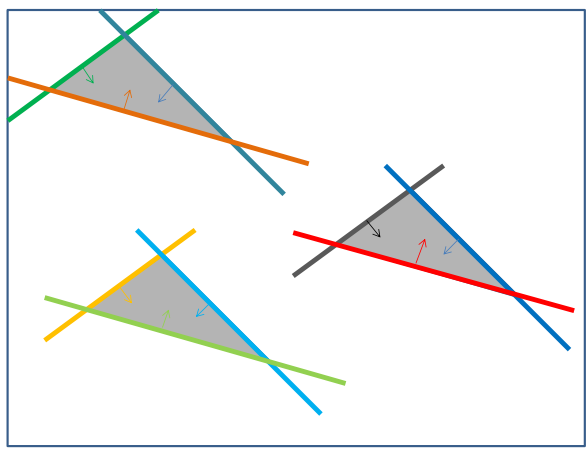

Fig. 2: Several convex polytopes defined by the intersection of the positive half-spaces of linear cuts. The polytopes enclose the class of gray points, which is separated from the white background points.

A multiplayer perceptron with the architecture shown in Fig. 3 can handle this separation of input space. In this network, the first layer represents all the cuts of the input space necessary for classification. The colored triangles represent weighted connections. Unnecessary connections can have a weight of 0.

The binary outputs of the first layer indicate which side of each cut the input vector is on. These binary outputs can then be combined using any desired logical function. For the classification task, we only need to determine the combined intersection of several cuts for each island or cluster. The second layer of units accomplishes this. The first unit in the second layer computes the intersection of the positive half-spaces of the cuts needed to enclose Cluster 1. The second unit in that layer computes the conjunction of cuts needed to separate the second island and enclose Cluster 2. Given all the binary outputs of the second layer, we only need an output unit that fires if any output is a 1, corresponding to the logical function OR.

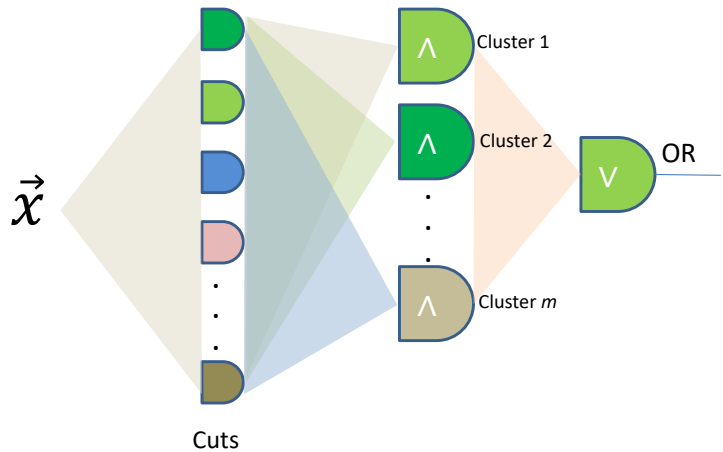

Fig. 3: Multiplayer perceptron implementing a conjunctive normal form of Boolean outputs, each one representing a cut of input space.

There are a few details to note: We do not reuse cuts for separate clusters. If a cut is needed twice, two separate gates are used in the first layer of cuts. If a cut does not separate space so that the grey points are on its positive half-space, we simply flip the signs of its weights. This is why we do not reuse cuts; sometimes, one or the other half-space must be the positive one. These complications can be avoided by using bipolar vectors with +1 or -1 components instead of binary components, because we can negate any output by changing the attached weight's sign. Here, we avoid such complications by not reusing cuts for different clusters.

Although the diagrams above show continuous regions of space, readers should keep in mind that the two classes are defined by discrete sets of points that can be separated by enclosing subsets within convex polytopes. In the limit, each point of one class could be enclosed. While that would result in an expensive network, here we are dealing with a proof of existence.

## 2    Transforming the neural network

The above network is structured as a disjunction of conjunctions. First, we prove that we can implement a disjunction of cuts using sequential gates per cut. Suppose we want to enclose the island shown in Fig. 4. We can use cuts 1 through 4, where the disjunction (union) of their positive regions is represented by the yellow region. A given point satisfies a cut if it is located in its positive half-space. In this case, it automatically belongs to the yellow class. Therefore, we can compute the cuts sequentially with perceptrons. When one of them outputs a 1, the entire computation is considered true. This strategy is illustrated in Fig. 5, which depicts the cascade of perceptrons recognizing points within the gray region.

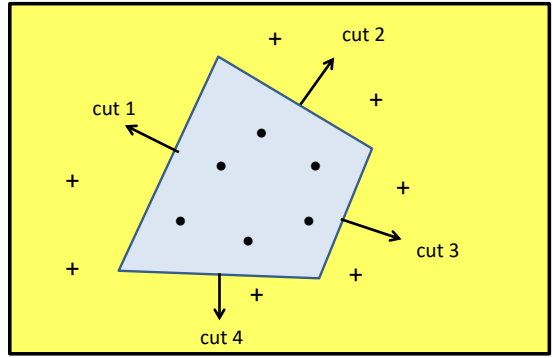

Fig. 4: Convex polytope enclosing the grey region. The union of positive half-spaces of the four cuts represents the yellow region.

Consider the sequence of gates shown in the upper diagram of Figure 5. Consider the sequence of gates shown in the upper diagram of Figure 5. Each gate represents a cut. Each gate receives the original input, the vector $x$, through skip connections, as well as the output of the previous gate. The weight S represents a large number. If we limit the vector $x$ to a maximum length of L and normalize the weight vector of each perceptron (normalization to $|w| = 1$ does not affect the cut represented by the perceptron), the inner product of $x$ and $w$ will have a maximum absolute value of L. In that case we pick S just much bigger than L. But then $x^T w + a \cdot S \geq 0$, if the output $a$ of the previous gate is 1, otherwise, if $a = 0$, the normal test is computed at each perceptron. This means that if one of the gates in the chain outputs a 1, then all subsequent gates will output a 1, regardless of the skipped input $x$. Then, the final output of the chain is also a 1. This represents the disjunction of logical decisions at each gate, which is necessary for detecting the yellow region in Fig. 4.

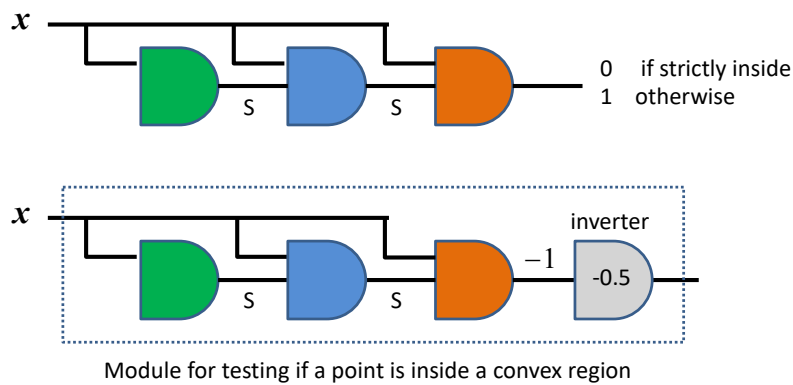

Module for testing if a point is inside a convex region

Fig. 5: Chain of perceptrons with skip connections. The upper diagram shows a disjunction of cuts. The lower diagram, with its output inverted, represents a conjunction of negated cuts.

Now, we can negate the output of the chain of perceptrons using an inverter implemented by a single perceptron, as shown in Fig. 5 (lower diagram). According to the De Morgan laws, the chain of gates now computes the negation of the disjunction of cuts, which represents the conjunction of the negated (inverted) cuts. The circuit produces a 1 if the point is inside the convex region defined by the gates' cuts and a 0 otherwise (see the gray region in Fig. 4). This is exactly what we need to replace the set of gates in the neural network of Fig. 3, which processes cluster 1. The same can be done for cluster 2 and so on. Then, we arrange each module representing a conjunction of cuts one after the other as shown in Fig. 6. As we saw before, this effectively represents a disjunction of modules, i.e., a disjunction of conjunctions of cuts, as in the neural network of Fig. 3.

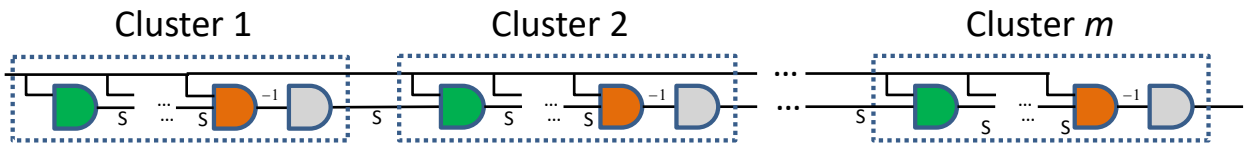

Fig. 6: Network equivalent to the multilayer perceptron in Fig. 3. Each module computes a conjunction of cuts. The arrangement of modules computes a disjunction of conjunctions of cuts.

## 3 Discussion

In this paper, we provide an intuitive description of three-layer neural networks: one layer of cuts, one layer of conjunctions of cuts, and one layer of disjunctions. After the layer of cuts, any Boolean function that can be implemented by the subsequent layers can be reduced to the disjunctive normal form shown in Figure 3. Regardless of how many layers the original network has, it can be reduced to this form.

The paper's main contribution is demonstrating that a network of this type can be transformed into a network with a single gate per layer and skip connections. Each layer forwards the initial input (the components of the input vector) and a single bit that tells the next layers, "This point belongs to a cluster of class 1." The proof of equivalence of the deep one-gate-per-layer network to the disjunctive normal form network is simpler than the proof in [1]. The visualizations complement other attempts at understanding the geometry of neural networks [3].

This result is primarily of theoretical interest.

**References**

[1] R. Rojas, "Deepest Neural Networks", arXiv: 1707.0261, 2017.
[2] R. Rojas, *Neural Networks*, Springer-Verlag, Berlin, 1996.
[3] J.Ch. Ye, Geometry of Deep Learning, Springer Verlag, Singapore, 2022.